\documentclass[letterpaper, 10 pt, conference]{ieeeconf}  % Comment this line out if you need a4paper

\IEEEoverridecommandlockouts                              % This command is only needed if 
\usepackage{graphicx}
\usepackage{xcolor}
\usepackage{hyperref}
\usepackage{booktabs}
\usepackage{amssymb}
\usepackage{amsmath}
\usepackage{nomencl}
\usepackage{gensymb}
\usepackage{siunitx}
\usepackage{subfig}
\usepackage{dblfloatfix} 
\usepackage[ruled,vlined]{algorithm2e}

\title{\LARGE \bf
InstructRobot: A Model-Free Framework for Mapping Natural Language Instructions into Robot Motion
}

\author{
Iury Cleveston$^{1}$, Alana C. Santana$^{2}$, Paula D. P. Costa$^{3}$, Ricardo R. Gudwin$^{4}$, \\ Alexandre S. Simões$^{5}$, Esther L. Colombini$^{6}$% <-this % stops a space
\thanks{$^{1}$Ms. Iury Cleveston. Institute of Computing, Hub of Artificial Intelligence and Cognitive Architectures (H.IAAC), University of Campinas. 
        {\tt\small iury.cleveston@ic.unicamp.br}}%
\thanks{$^{2}$Ms. Alana C. Santana. Institute of Computing, Hub of Artificial Intelligence and Cognitive Architectures (H.IAAC), University of Campinas. 
        {\tt\small alana.santana@ic.unicamp.br}}%
\thanks{$^{3}$Dr. Paula D. P. Costa. Faculty of Electrical and Computer Engineering, Hub of Artificial Intelligence and Cognitive Architectures (H.IAAC), University of Campinas.
        {\tt\small paulad@unicamp.br}}%
\thanks{$^{4}$Dr. Ricardo R. Gudwin. Faculty of Electrical and Computer Engineering, Hub of Artificial Intelligence and Cognitive Architectures (H.IAAC), University of Campinas. 
        {\tt\small gudwin@unicamp.br}}%
\thanks{$^{5}$Dr. Alexandre S. Simões. Hub of Artificial Intelligence and Cognitive Architectures (H.IAAC), São Paulo State University.
        {\tt\small alexandre.simoes@unesp.br}}%
\thanks{$^{6}$Dr. Esther L. Colombini. Institute of Computing, Hub of Artificial Intelligence and Cognitive Architectures (H.IAAC), University of Campinas. 
        {\tt\small esther@ic.unicamp.br}}%
}

\begin{document}

\maketitle
\thispagestyle{empty}
\pagestyle{empty}

%%%%%%%%%%%%%%%%%%%%%%%%%%%%%%%%%%%%%%%%%%%%%%%%%%%%%%%%%%%%%%%%%%%%%%%%%%%%%%%%
\begin{abstract}

The ability to communicate with robots using natural language is a significant step forward in human-robot interaction. However, accurately translating verbal commands into physical actions is promising, but still presents challenges. Current approaches require large datasets to train the models and are limited to robots with a maximum of 6 degrees of freedom. To address these issues, we propose a framework called InstructRobot that maps natural language instructions into robot motion without requiring the construction of large datasets or prior knowledge of the robot's kinematics model. InstructRobot employs a reinforcement learning algorithm that enables joint learning of language representations and inverse kinematics model, simplifying the entire learning process. The proposed framework is validated using a complex robot with 26 revolute joints in object manipulation tasks, demonstrating its robustness and adaptability in realistic environments. The framework can be applied to any task or domain where datasets are scarce and difficult to create, making it an intuitive and accessible solution to the challenges of training robots using linguistic communication. Open source code for the InstructRobot framework and experiments can be accessed at \url{https://github.com/icleveston/InstructRobot}.

\end{abstract}

%%%%%%%%%%%%%%%%%%%%%%%%%%%%%%%%%%%%%%%%%%%%%%%%%%%%%%%%%%%%%%%%%%%%%%%%%%%%%%%%
\section{INTRODUCTION}

The ability to communicate effectively with robots is essential for completing complex tasks. Accurately translating verbal commands into physical actions is a promising task~\cite{zhu2021deep,zhang2015towards}, but still presents challenges~\cite{sheridan2016human,bonarini2020communication,broadbent2017interactions}. They often come with a high cost in terms of training, requiring vast datasets to train the models. Furthermore, jointly learning language and motion only works for robots with a maximum of 6 degrees of freedom (DoF), where the inverse kinematics model is already known~\cite{jiang2022vima,stepputtis2020language,finn2017one}. These limitations hinder the progress of the field, making it impractical to learn, for complex robots, robot motion without a prior kinematics model and limiting the use of natural language to manipulate robots in everyday life.

\begin{figure}[t]
    \centering
    \includegraphics[width=0.45\textwidth]{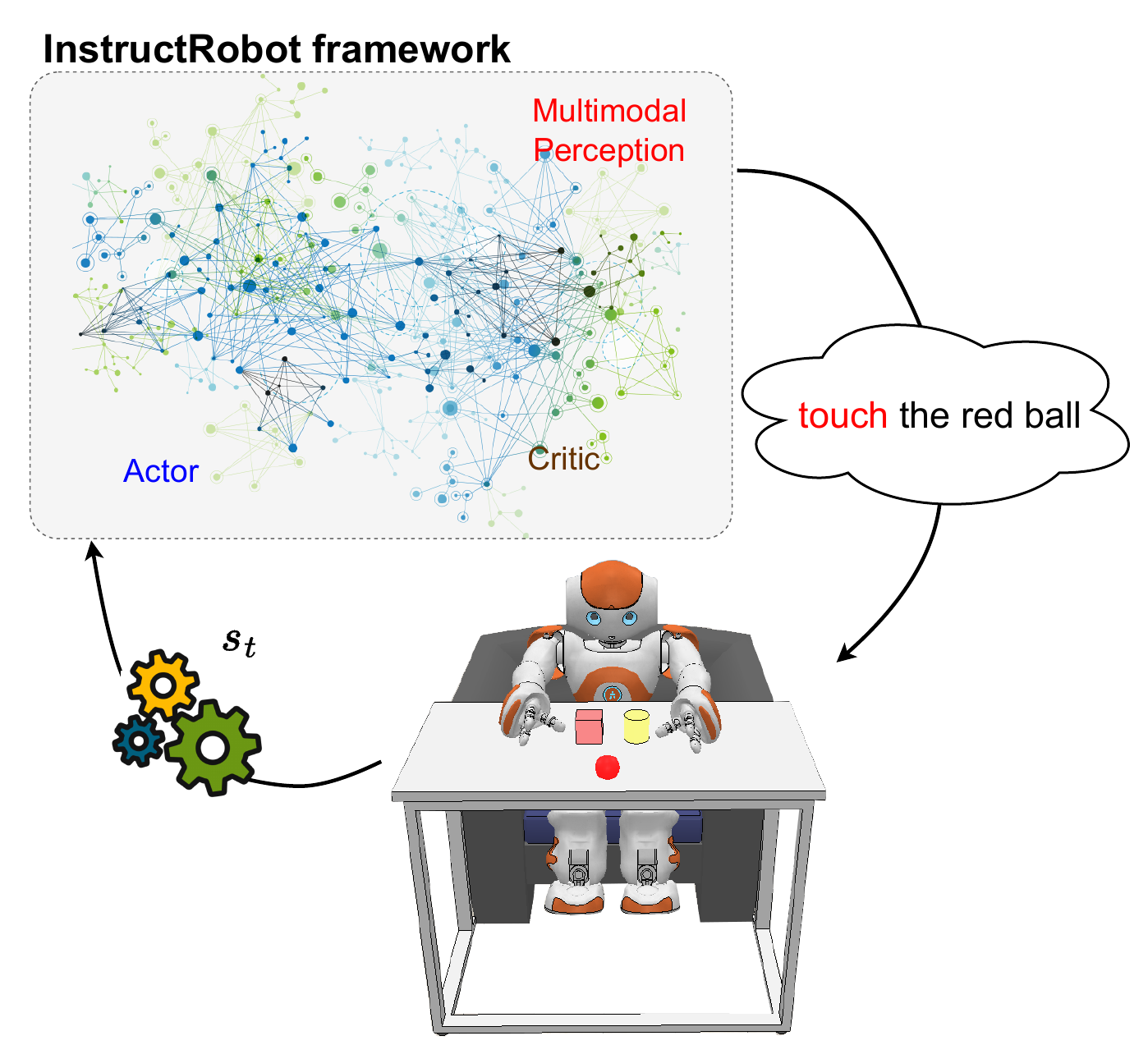}\hfill
    \caption{An illustration of the InstructRobot framework. Our framework enables natural language instructions to be mapped into robot motion without requiring the creation of datasets or knowledge of inverse kinematic models.}
    \label{fig:framework_proposto_1}
\end{figure}

In this work, we present InstructRobot, a framework that maps natural language instructions into robot motion without requiring the construction of large datasets or prior knowledge of the robot's inverse kinematics model. InstructRobot employs a reinforcement learning (RL) algorithm that is highly adaptable and easy to implement, making supervised training unnecessary. Our approach simplifies the learning process and offers an intuitive and accessible solution to the challenges of training robots using linguistic communication. Our experiments demonstrate the joint learning of robot motion and language representations, a significant advancement in simplifying the training process. In addition, the framework's modular approach enhances its flexibility, allowing the inclusion of new perceptual modules and efficient adaptations to various tasks and domains.

To validate the effectiveness of the proposed framework (Figure~\ref{fig:framework_proposto_1}), this work employs the NAO robot~\cite{6190579} in object manipulation tasks. NAO is an extremely complex robot with 26 revolute joints that must be controlled in a highly realistic environment. This deliberate choice highlights the robustness and adaptability of the proposed framework, demonstrating its ability to deal with highly demanding environments and tasks. InstructRobot allows multi-modality of input data, and any architecture or reinforcement learning algorithm can be used to formulate the agent. The fundamental aspect of our framework is to assemble an environment in such a way the agent is encouraged to explore and learn an RL policy that better maps natural language instructions into robotic motion. Searching for an RL policy by exploring the environment avoids the task generalization issues frequently present when dataset construction is needed~\cite{jiang2022vima}. The framework can be applied to any task or domain, especially where datasets are scarce and difficult to create.

Therefore, the main contributions of our work are:

\begin{enumerate}
    \item To the best of our knowledge, InstructRobot is the first instructional framework that learns language representations and model-free kinematics motion simultaneously;

    \item  To the best of our knowledge, this is the first work in the literature to validate an instructional learning approach using a complex 26-revolute joint robot like NAO;
    
    \item  To the best of our knowledge, this is the first work that uses reinforcement learning to map instructions to motion, eliminating the need for large datasets;
    
    \item Validation of the framework's effectiveness through challenging object manipulation experiments using state-of-the-art neural networks trained via reinforcement learning;

    \item  Open source code for the InstructRobot framework and experiments can be accessed at \url{https://github.com/icleveston/InstructRobot}.

\end{enumerate}

\section{RELATED WORK}

Recent studies have examined the idea of teaching robots through natural language. One approach was proposed by Ahn et al.~\cite{ahn2022visually} and emphasizes a robot's need to refer to its task history, particularly when executing a series of pick-and-place manipulations through instructions given sequentially. The authors created a history-dependent dataset with pick-and-place operations instructed through natural language. A human provides the robot with linguistic instructions to move certain blocks individually to build a specific structure. The robot's main objective is to estimate the target block's position before and after executing the instructions. The dataset provides synthetic RGB images of the workspace, a set of natural language instructions written by six annotators, the objects' bounding boxes, and heat maps showing the target object. Six annotators describe the same operation for each task, in their own style. The model is trained via supervised learning, using the heatmap as ground truth.

Kuo et al.~\cite{kuo2022trajectory} conducted a study on trajectory prediction systems that utilize linguistic representations, where the model is trained using sample trajectories with labels. However, the study highlights the need for motion prediction datasets with natural language annotation. To address this limitation, the authors expanded the Argoverse~\cite{chang2019argoverse} dataset with synthetic language descriptions, which were limited by the types of filters used to generate these descriptions. They also sampled 40,000 trajectories from the Waymo~\cite{sun2020scalability} dataset and employed human labor to annotate their labels. The authors observed that annotated sentences had a much more diverse vocabulary than synthetic language, but the annotation process was quite laborious. 

Due to the limitations of the available datasets, some studies have been restricted to examining how robots understand natural language through tasks that do not require physical actions in the real world. These tasks can be performed using conventional datasets consisting of pairs of image-instruction previously developed in the literature. For instance, Cheang et al.~\cite{cheang2022learning} focused on inferring the objects' category and estimating the 6-DoF information for invisible objects of known classes from natural language instructions provided to a robot. They used RefCOCO~\cite{kazemzadeh2014referitgame} as a benchmark because it is a more established benchmark in the literature, and existing ground truth is available. Additionally, Yang et al.~\cite{yang2022interactive} explored object attributes in disambiguation and developed an interactive understanding system that can resolve ambiguities through dialogues.

Various robot manipulation tasks require different skills and formats, including following instructions~\cite{stepputtis2020language}, one-shot imitation~\cite{finn2017one}, rearrangement~\cite{batra2020rearrangement}, constraint satisfaction~\cite{brunke2022safe}, and reasoning~\cite{shridhar2020alfred}. Multiple physics simulation benchmarks have been introduced to study these tasks. For instance, iGibson~\cite{shen2021igibson, li2021igibson, srivastava2022behavior, li2023behavior} simulates interactive household scenarios, Ravens~\cite{zeng2021transporter} rearranges deep features to infer spatial displacements from visual input and exhibit superior sample efficiency on several tabletop manipulation tasks. Robosuite~\cite{zhu2020robosuite} is a modular simulation framework and benchmark for robot learning that aims to facilitate research and development of data-driven robotic algorithms and techniques. CALVIN~\cite{mees2022calvin} develops long-horizon language-conditioned tasks, Meta-World~\cite{yu2020meta} is a benchmark of simulated manipulation tasks with everyday objects contained in a shared, tabletop environment. AI2-THOR~\cite{ehsani2021manipulathor} is a framework that supports visual object manipulation and procedural generation of environments.

\begin{figure*}[b!]
\centering
    \includegraphics[width=\textwidth]{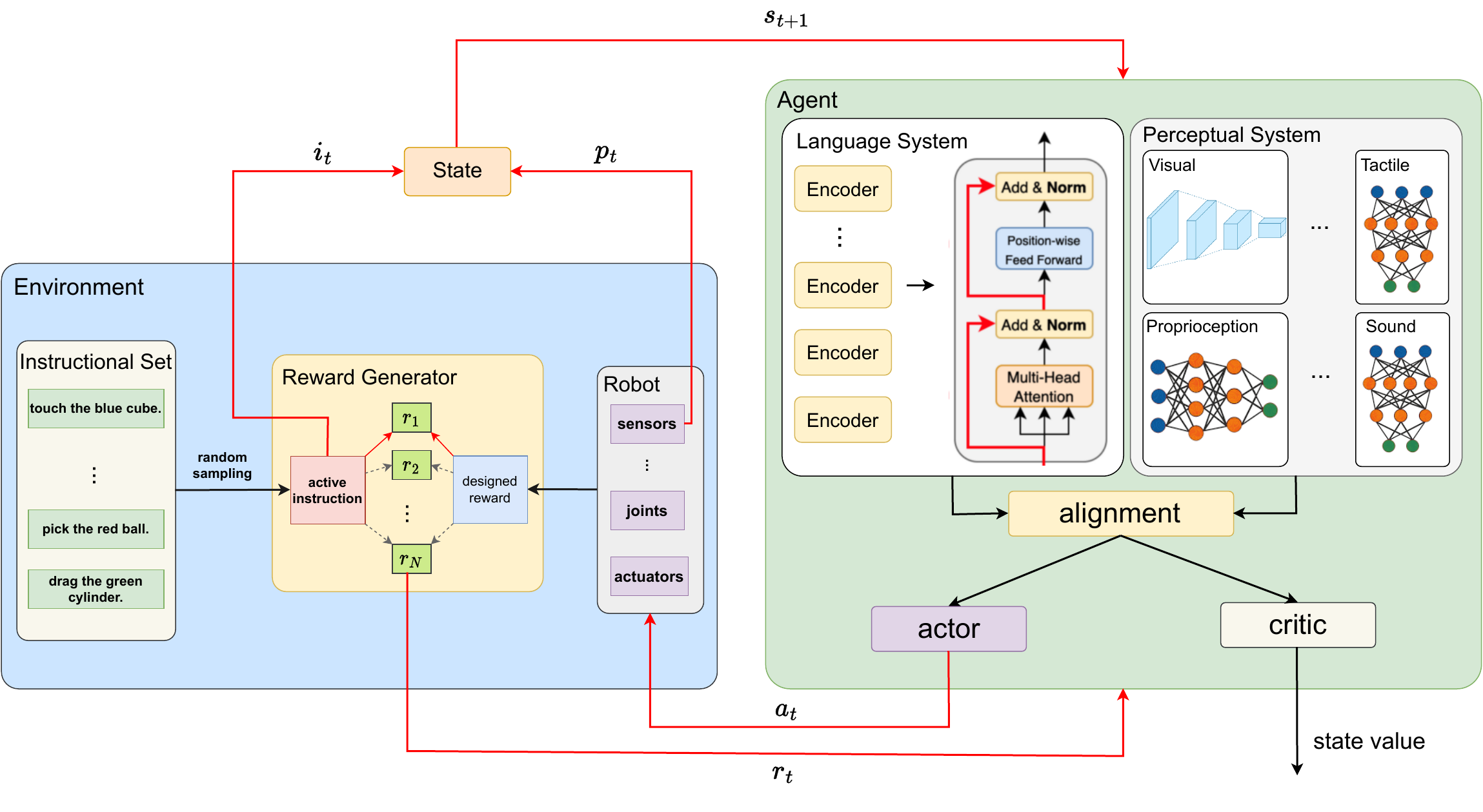}\hfill
    \caption{The InstructRobot framework comprises two main blocks: environment and agent. The Environment block was designed to simplify the process of generating task instructions and rewards and is composed mainly of the Instructional Set, Reward Generator, and Robot modules. For every episode, an instruction $i_t$ is randomly selected from the Instructional Set and becomes the active instruction of the environment, providing a designed reward function to evaluate the robot's actions. The Agent block comprises the Language System, Perceptual System, Alignment, and Actor and Critic modules. The agent receives a state $\textbf{s}_{t+1}$ that comprises the active instruction $i_t$ and the perceptual information $\textbf{p}_t$ from the environment, processes it in its subsystems and acts in the environment by sending an action $\textbf{a}_t$, receiving a reward $r_t$. In this process, the agent learns a policy $\pi(\textbf{a}|\textbf{s})$ that maps the instruction into robot motion.}
    \label{fig:all_framework}
\end{figure*}

A significant challenge for approaches that require annotated data is dealing with generalization. VIMA~\cite{jiang2022vima} pursued an imitation learning strategy using Transformer architectures and multimodal prompts to train a robotic manipulator with previously learned inverse kinematics model in several pick-and-place and wipe tasks. To train the model and improve generalization, they generated a large offline dataset containing 50k trajectories per task, totaling 650k trajectories. An expert oracle collected the trajectories for all tasks, and a behavior clone algorithm was employed to learn the pick-and-place actions. Despite facilitating data sampling, this approach still has several disadvantages, such as the low generalization that imitation learning methods present in general and that the learning of complex tasks is dependent on the oracle. Our framework is related to these works, mainly in object manipulation tasks through natural language instructions. However, our framework stands out for tackling the challenge of the joint learning kinematics model and language representation while avoiding the creation of datasets.

\section{OUR FRAMEWORK}

The InstructRobot framework comprises two main blocks: environment and agent, as illustrated in Figure~\ref{fig:all_framework}. Each block has specialized modules that perform different tasks. We designed our framework to support  various sensory modalities, robots, reinforcement learning algorithms, and neural network models. Additionally, our framework is highly modular, making it easy to modify and add new modules and features.

\subsection{Environment}

The Environment block was designed to simplify the task instructions and rewards generation. The Instructional Set and Reward Generator modules eliminate the need for creating costly datasets for supervised or imitation learning. The Instructional Set contains the instructions required to train the agent. These instructions can be easily created through artificialization or captured from other data sources. A corresponding reward function is designed to encourage the robot to execute a specific instruction successfully. The designed reward submodule provides a reward function for each instruction in the Instructional Set. During training, instructions are randomly sampled from the Instructional Set, and the designed reward function associates the sampled instruction with its respective reward function $\left ( i_{t}, r_{t}\right )$. Finally, the Robot module contains all the physical characteristics of the actuators and sensors of the robot used. This module provides all the perceptual information $\textbf{p}_t$ necessary to create the state. Once the robot's physical characteristics are changed, only the Robot module needs changing in the framework, making it easy to adapt the system to different robots and tasks. At each time step $t$, the environment sends to the agent information about its next state $\textbf{s}_{t+1}$, the reward value $r_t$ achieved by the robot when executing the instruction $i_t$, and whether the episode is complete.

The state $\textbf{s}_t$ is composed of three consecutive observations of the scene as $\textbf{s}_t = [\textbf{o}_{t}, \textbf{o}_{t-1}, \textbf{o}_{t-2}]$. Stacking observations to compose a state is mainly helpful in avoiding perceptual aliasing, where multiple states may give rise to the same perception. Each observation $\textbf{o}_t$ is composed of the active instruction and the robot's perceptual information as $\textbf{o}_t = \mathrm{[i_{t}, \textbf{p}_{t}]}$. Our state utilizes a multi-modal observational space and can be extended to include any perceptual or linguistic information, such as natural language instructions, images, proprioception, tactile information, and others. 

\subsection{Agent}

The Agent block receives the state $\textbf{s}_{t}$ from the environment and selects an action $\textbf{a}_t$ that maximizes the episodic accumulated return over time. To achieve this, the agent uses different modules to process state information. The Language System is responsible for processing all linguistic information received, while the Perceptual System has various sensory submodules, each responsible for processing information from a particular sensory modality. At this stage, sensory information is processed in isolation, with a NN to each sensor modality. Figure~\ref{fig:all_framework} exemplifies some sensory submodules that can be used, but others can be added easily. The Alignment module aligns the perceptual and linguistic information to create a condensed representation that guides the actor and critic modules -- traditional parts of reinforcement learning algorithms. The actor module generates the most appropriate action $\textbf{a}_{t}$, sending it to the environment to execute. In contrast, the critic module generates the state value for optimizing the RL algorithms. The Language System and the Perceptual System's submodules can be represented by any system, from pre-trained neural networks to even hard-coded feature extractors. Moreover, the framework allows different RL algorithms for end-to-end optimization of all agent parameters or just some modules.

\subsection{Problem Definition}

To evaluate our framework, we train a policy $\pi\left ( \textbf{a}_{t+1} | \textbf{s}_{t} \right )$ for mapping instructions into robot motion conditioned on a state $\textbf{s}_{t}$, composed of three observations containing a natural language instruction $i_{t}$, visual data $\left \{ \textbf{x}_{j} \right \}_{j=t-2}^{t}$, proprioception  data $\left \{ \textbf{y}_{j} \right \}_{j=t-2}^{t}$, and tactile data $\left \{ \textbf{l}_{j} \right \}_{j=t-2}^{t}$, where \textit{t} is the current time step. The natural language instruction $i_{t}$ consists of sentences to manipulate the objects, with a vocabulary size of 10. The visual observations consist of RGB images from \textit{K} cameras, with each $x_{j}$ of size $H \times W \times 3$ (512 height, 256 width, 3 channels). Following \cite{jiang2022vima}, we use $K = 2$ with cameras on the front and top positions concerning the robot. The proprioception $\textbf{y}_{j}$ $\in$ $\mathbb{R}^{d_{prop}}$ consists in 26 angular joint poses. The tactile data $\textbf{l}_{j}$ $\in$ $\mathbb{R}^{d_{tactil}}$ consists of 6 binary values that indicate collisions between objects and the robot's fingers. The action space $\textbf{a}_{t}$ consists of 26 angular joint poses. We do not employ previous inverse kinematic models in our experiments.
 
\textbf{Feature Encoding}. We implement state-of-the-art neural networks to process linguistic and perceptual state information. In the Language System, we use the Neural Transformer~\cite{vaswani2023attentionneed} to process the instructions, with three layers, two heads per layer, and 50 hidden units. Given the instruction $i_{t}$, the transformer obtains a contextualized representation $\textbf{G}_{t} = \mathrm{Encoder_3}(i_t)$ conditioned on the encoded instruction. This enables learning relationships among the different tokens in the instruction. We use the transformer’s encoders with a self-attention mechanism to learn such relationships as
\begin{equation}
    \mathrm{Attn}(\textbf{Q},\textbf{K},\textbf{V}) = \mathrm{Softmax}\Big(\frac{\textbf{W}_{Q}\textbf{Q}(\textbf{W}_{K}\textbf{K})^{T}}{\sqrt{d}}\Big)\textbf{W}_{V}\textbf{V,}
    \label{eq1:self_attention}
\end{equation}

\noindent where $\textbf{W}_{Q}$, $\textbf{W}_{K}$, $\textbf{W}_{V}$ are learnable parameters. 

In the Perceptual System, we activate the visual, proprioception, and tactile modules. The visual observations are processed by two convolutional layers with 12 input channels, two max pooling functions with kernel and stride $2 \times 2$, and two linear layers with 256 units at the output, generating the embedding $\textbf{V}_{t}$. Proprioception and tactile information are processed together by a network with a linear layer of 128 units generating the $\textbf{H}_{t}$ embedding.

\textbf{Action Prediction}. We concatenate the output embeddings $\textbf{G}_{t}$, $\textbf{V}_{t}$, and $\textbf{H}_{t}$ in a single vector $\textbf{B}_{t}$. We employed three linear layers with 500, 256, and 128 units in the actor module to generate the action prediction $\textbf{a}_{t}$ $\in$ $\mathbb{R}^{d_{prop}}$ from $\textbf{B}_{t}$. We use \textit{tanh} activation function between each layer. Finally, the actor's output is used as the mean value to initialize a multivariate normal distribution. 

\textbf{Training and Inference}. We train all modules using the Proximal Policy Optimization (PPO)~\cite{schulman2017proximal} algorithm. PPO is a robust optimization algorithm with good convergence that requires little hyperparameter tuning. Reward normalization was employed to reduce variance, but we do not utilize Generalized Advantage Estimate (GAE)~\cite{schulman2018highdimensional} at this stage. We train all modules from scratch using a reinforcement learning loss to maximize the episodic accumulated return by the agent throughout the training episodes, as follows

\begin{equation}
J(\boldsymbol\theta) = \mathbb{E}_t\Big[\frac{\pi_{\boldsymbol\theta}(a|s)}{\pi_{\boldsymbol\theta_{old}}(a|s)} \textbf{A}_t\Big],
\label{eq:ppo_J}
\end{equation}

\noindent where $\textbf{A}_t$ is the advantage function and $\boldsymbol\theta_{old}$ are the old policy's parameters. 

Finally, to generate the state value used in the optimization process, we use a network with three linear layers in the critic module with 500, 256, and 128 units, respectively, and \textit{tanh} activation function between each layer.

\section{Experiments}

We designed three experiments to evaluate the efficacy
of our framework. We incrementally modified the environment to increase the baseline agent's complexity to complete the object manipulation task. We control NAO arms, hands, and fingers; the legs, torso, and head are fixed in default positions. Moreover, each joint has a maximum and minimum angular amplitude allowed to operate. Collisions are detected between the fingers and the object being simulated in the scene. To evaluate the agent's performance, we use the mean episodic accumulated reward. 

\textbf{Hardware and Software Configurations.} We implemented our method with PyTorch 1.3.1 and CUDA v12.2. We conducted experiments on Nvidia RTX 4090 with 24Gb, motherboard Asus Rog Strix Z790-A Gaming, Intel Core i7-13700KF CPU @ 5.4GHz, RAM Corsair DDR4 1x32 Gb @ 2666MHz, disk Seagate Barracuda 2Tb, and Operating System Ubuntu v22.04. 

\subsection{Experiment I - Single-Instruction}

\begin{figure}[t!]
    \centering
    \includegraphics[width=0.48\textwidth]{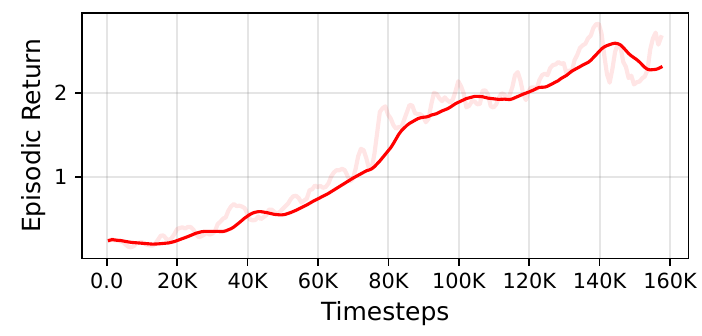}
    \caption{Mean episodic return for our single-instruction experiment. The agent could learn a policy to complete the task, achieving a mean return of $2.4$.}
    \label{fig:baseline_reward}
\end{figure}

\begin{figure}[b!]
    \centering
    
    \subfloat[\centering Frame 0]{{\includegraphics[width=3.35cm]{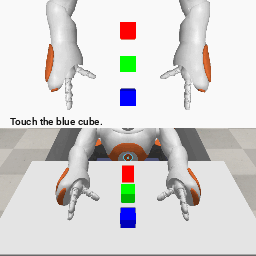} }}
    \hspace{0em}
    \subfloat[\centering Frame 31]{{\includegraphics[width=3.35cm]{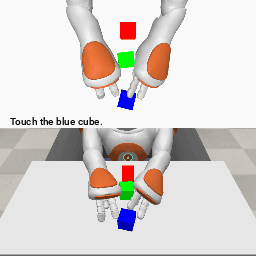} }}
    \hspace{0em}
    
    \caption{Policy execution for our single-instruction experiment. The agent's goal was to touch only the blue cube. Note that the agent can move its arms and fingers to the correct cube and keep them in that position for the entire trajectory.}
    \label{fig:exp_1}
\end{figure}

Our aim with this experiment is to prove that an agent can learn the inverse kinematics model by receiving a reward signal associated with language instruction, even when environmental distractions exist. To test this, we set up an environment with three differently colored cubes arranged in line with only one instruction, ``\textit{Touch the blue cube.}'', in the Instructional Set, as shown in Figure~\ref{fig:exp_1}. We configured the Reward Generator to reward $1.0$ whenever the agent touches the blue cube or $0.0$ if it touches any other object or does not touch the blue cube. We used a multi-modal observational space consisting of natural language instructions and images. During training, we collected 24 roll-outs with a trajectory length of 32 steps. We trained our agent using a multivariate normal distribution, with a standard deviation 0.36. We opted for a low learning rate of $\num{1e-5}$ and refined the policy for 60 epochs to account for our small batch size. The baseline agent was trained for $\num{15e4}$ steps and the instruction given to our agent during training remained the same in all training steps.

The results presented in Figure~\ref{fig:baseline_reward} indicate that the agent's mean return value per episode gradually increased over time, starting from zero in the initial steps and reaching a peak of 3.1 at the end of the training phase. The mean reward was 2.4, demonstrating that the agent learned to follow instructions correctly during training. The policy execution showed it could touch the blue cube and maintain its joints in the same position throughout the episode. The learned policy avoided unnecessary joint motion while performing the task.

\subsection{Experiment II - Multi-Instructions}

Our second experiment evaluated whether the agent could learn to follow multiple instructions in a scene by using rewards as guidance. To do this, we kept the baseline settings of our agent from the first experiment while adding three new instructions to the Instructional Set: 1) ``\textit{Touch the blue cube.}'', 2) ``\textit{Touch the red cube.}'', and 3) ``\textit{Touch the green cube.}''. Additionally, we modified the Reward Generator to create new reward functions that matched the new instructions. The Reward Generator was set up to reward 1.0 if the agent touched the correct cube and 0.0 if it touched a wrong cube or failed to touch any cube.

\begin{figure}[ht!]
    \centering
    \includegraphics[width=0.48\textwidth]{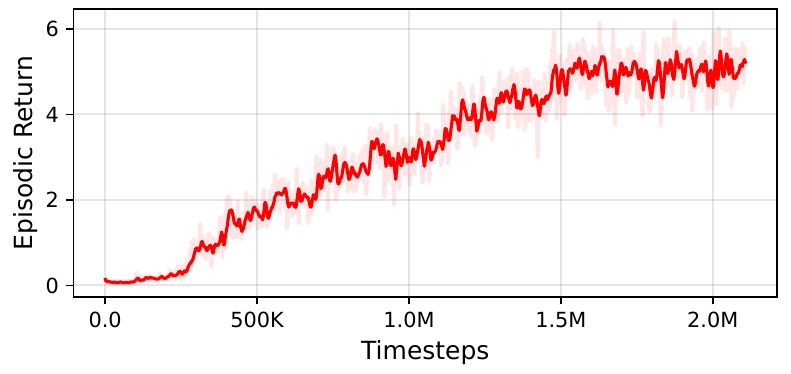}
    \caption{Mean episodic return for the multi-instructions experiment. The agent was trained for 2.2 million steps, achieving a mean episodic return of 5.}
    \label{fig:exp_2_reward}
\end{figure}

\begin{figure}[ht!]
    \centering
    
    \subfloat[\centering Frame 0]{{\includegraphics[width=3.35cm]{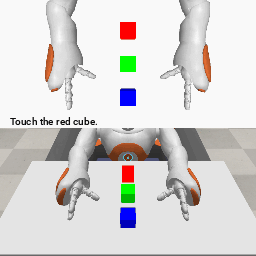} }}
    \hspace{0em}
    \subfloat[\centering Frame 31]{{\includegraphics[width=3.35cm]{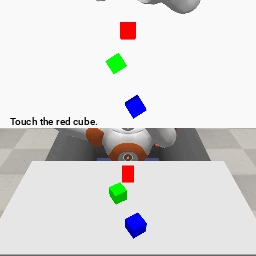} }}
    \hspace{0em}
    
    \caption{Policy execution for our multi-instructions experiment. The goal was to touch the red cube, but the agent could not follow the active instruction appropriately.}
    \label{fig:exp_2_red}
\end{figure}

The agent was trained for 2.2 million steps, and it was able to achieve a good policy for most tasks with a mean episodic reward of 5, as depicted in Figure~\ref{fig:exp_2_reward}. The agent could consistently touch the blue and green cubes during the experiment, proving that it could associate instructions with colors without defining specific rewards for this. However, it exhibited excessive joint motion that was not optimal for the task, often touching multiple cubes. Unfortunately, the agent could not complete the task of touching the red cube, as shown in Figure~\ref{fig:exp_2_red}. We hypothesized that the difficulty in learning this instruction might be due to the arrangement of objects on the table, where the red cube is placed at the end. We think a different arrangement of objects on the table may help the agent touch the red cube during the exploration stage. Moreover, we believe that including proprioception in the state representation may help improve exploration and learning of different tasks.

\subsection{Experiment III - Improving Multi-Instructions}

\begin{figure}[b!]
    \centering
    \includegraphics[width=0.48\textwidth]{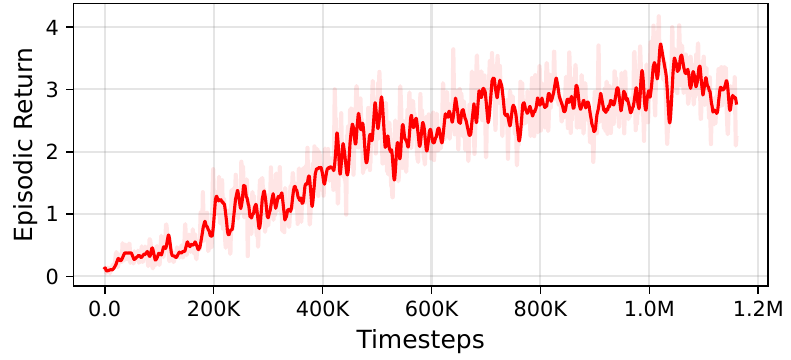}
    \caption{Mean episodic return for the improving multi-instructions experiment. The agent achieved a mean episodic return of 3.1.}
    \label{fig:exp_3_reward}
\end{figure}

In the third experiment, our goal was to modify the scene setup of the multi-instructions experiment to improve the agent's exploration. We kept the new environment similar to the previous ones but included the agent's joint position as the proprioception information to the state representation. Regarding the scene modifications, we set up the same three cubes but positioned triangularly. We also have slightly modified the reward system; now, we consider the number of fingers that touch the object and reward the agent accordingly. Whenever the agent successfully executes the active instruction and touches the correct cube, we reward them with a score of $1.0$ for each finger that touches the cube. However, if the agent touches the wrong cube, we apply a penalty of $-0.1$ for each finger involved. Therefore, the maximum reward is 6 when the agent keeps its six fingers touching the proper cube for the entire episode.

\begin{figure}[ht!]
    \centering
    
    \subfloat[\centering Frame 0]{{\includegraphics[width=3.35cm]{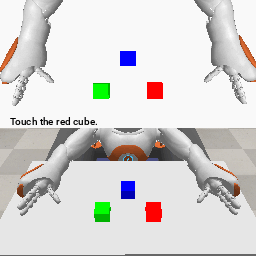} }}
    \hspace{0em}
    \subfloat[\centering Frame 31]{{\includegraphics[width=3.35cm]{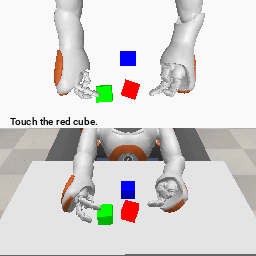} }}
    \hspace{0em}
    
    \caption{Policy execution for our improving multi-instructions experiment. The goal was also to touch the red cube and the agent could execute the instruction appropriately.}
    \label{fig:exp_3_red}
\end{figure}

The new agent learned a good policy, as demonstrated by the mean episodic return in Figure~\ref{fig:exp_3_reward}, which reached 3.1. This modification made the state more representative since the agent could better localize its joints concerning the objects. As shown in Figure~\ref {fig:exp_3_red}, the agent could correctly execute the desired instruction. The penalty given in the reward function when the agent touches the wrong cubes caused it to decrease the contact with these cubes. The novel scene with the objects positioned triangularly allowed the agent to explore the space better and touch all objects during training. Our modification to the agent did improve the final learned policy as the agent could reach more regions of the state space than in previous experiments. 

The experiments demonstrated that an agent can learn to follow natural language instructions in a complex, multimodal environment, by learning the ability to associate language instructions with actions based on reward signals. Improvements achieved by including proprioception and modifying the scene arrangement in later experiments demonstrated the importance of the modular framework nature that supports multimodal data.

\section{CONCLUSIONS}

In this study, we presented the InstructRobot framework, an innovative approach to translating natural language instructions into robotic motion without needing an inverse kinematics model or dataset creation. Our experiments show the feasibility of learning the kinematics model of a complex robot through environmental exploration, paving the way for the development of robust reinforcement learning policies that associate instructions with appropriate robot motion. Additionally, our approach's flexibility allows for several interesting extensions. Notably, exciting questions for future research include creating more automatic or jointly learned reward generators to minimize designer intervention and exploring the framework as a pre-training platform for basic tasks where reward design is less costly than the creation of datasets, significantly advancing results in the area. Furthermore, we aim to investigate whether our framework produces more promising generalization results than other training approaches and how the use of Large Language Models in the Language System can contribute to achieving promising generalization results.

\section*{ACKNOWLEDGMENT}

E. L. Colombini is partially funded by CNPq PQ-2 grant (315468/2021-1), A. S. Simões is partially funded by CNPq PQ-2 grant (312323/2022-0), A. C. Santana and I. Cleveston are funded by  MCTI/Softex. This project was supported by the Ministry of Science, Technology, and Innovation of Brazil, with resources granted by Federal Law 8.248 of October 23, 1991, under the PPI-Softex. The project was coordinated by Softex and published as Intelligent Agents for Mobile Platforms based on Cognitive Architecture technology [01245.013778/2020-21].

\bibliographystyle{Transactions-Bibliography/IEEEtran}
\bibliography{root.bib}
%%%%%%%%%%%%%%%%%%%%%%%%%%%%%%%%%%%%%%%%%%%%%%%%%%%%%%%%%%%%%%%%%%%%%%%%%%%%%%%%

\end{document}